\documentclass[10pt,twocolumn,letterpaper]{article}

\usepackage[accsupp]{axessibility} 
\usepackage{cvpr}              

\usepackage[dvipsnames]{xcolor}

\definecolor{cvprblue}{rgb}{0.21,0.49,0.74}
\usepackage[pagebackref,breaklinks,colorlinks,citecolor=cvprblue]{hyperref}
\usepackage{color}
\usepackage{multirow}
\usepackage{bbding}
\usepackage{xcolor}

\def\paperID{3027} 
\def\confName{CVPR}
\def\confYear{2024}

\title{SDSTrack: Self-Distillation Symmetric Adapter Learning for \\
Multi-Modal Visual Object Tracking}
\author{Xiaojun Hou$^1$, Jiazheng Xing$^1$, Yijie Qian$^1$, Yaowei Guo$^1$, Shuo Xin$^1$, Junhao Chen$^1$,\\
Kai Tang$^1$, Mengmeng Wang$^1$, Zhengkai Jiang$^2$, Liang Liu$^{3}$\footnotemark[1], Yong Liu$^{1}$\footnotemark[1]\\
$^1$Zhejiang University\quad$^2$Youtu Lab, Tencent\quad$^3$Huzhou Institute, Zhejiang University
\\
\texttt{\small $^1$\{xiaojunhou, jiazhengxing, yijieqian, yaoweiguo, shuoxin, chenjunhao}
\\
\texttt{\small kaitang, mengmengwang\}@zju.edu.cn}\quad\;\texttt{\small $^{1 *}$yongliu@iipc.zju.edu.cn} \\
\texttt{\small $^2$zhengkjiang@tencent.com}\quad\;\texttt{\small $^{3 *}$leonliuz@zju.edu.cn}
}

\begin{document}
\maketitle
\renewcommand{\thefootnote}{\fnsymbol{footnote}}
\footnotetext[1]{Corresponding authors.}
\begin{abstract}
Multimodal Visual Object Tracking (VOT) has recently gained significant attention due to its robustness. 
Early research focused on fully fine-tuning RGB-based trackers, which was inefficient and lacked generalized representation due to the scarcity of multimodal data. Therefore, recent studies have utilized prompt tuning to transfer pre-trained RGB-based trackers to multimodal data. However, the modality gap limits pre-trained knowledge recall, and the dominance of the RGB modality persists, preventing the full utilization of information from other modalities.
To address these issues, we propose a novel symmetric multimodal tracking framework called SDSTrack. We introduce lightweight adaptation for efficient fine-tuning, which directly transfers the feature extraction ability from RGB to other domains with a small number of trainable parameters and integrates multimodal features in a balanced, symmetric manner.
Furthermore, we design a complementary masked patch distillation strategy to enhance the robustness of trackers in complex environments, such as extreme weather, poor imaging, and sensor failure.
Extensive experiments demonstrate that SDSTrack outperforms state-of-the-art methods in various multimodal tracking scenarios, including RGB+Depth, RGB+Thermal, and RGB+Event tracking, and exhibits impressive results in extreme conditions. Our source code is available at: \url{https://github.com/hoqolo/SDSTrack}.
\end{abstract}
\vspace{-1em}    
\section{Introduction}
\label{sec:1_Introduction}
\begin{figure}[t]
  \centering
   \includegraphics[width=1.0\linewidth]{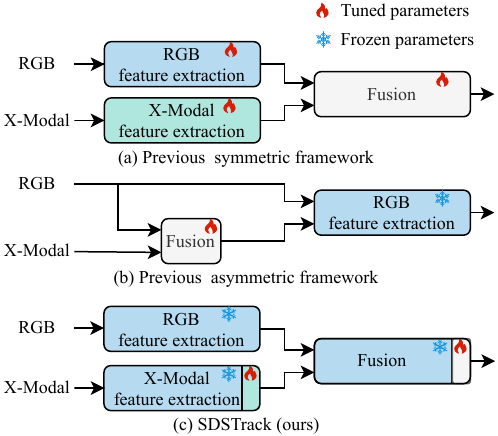}
   \caption{\textbf{Previous frameworks \emph{vs.} SDSTrack}.
   (a) Previous symmetric framework~\cite{yan2021depthtrack} has lots of training parameters and risk of overfitting. (b) Previous asymmetric framework~\cite{zhu2023ViPT} regards RGB as the primary modality and X-Modal as the auxiliary modality with prompt tuning. (c) Our proposed SDSTrack utilizes adapter-based tuning to fine-tune the pre-trained RGB-based tracker in a symmetric manner. ``X-Modal" denotes modalities other than RGB, which can be Depth, Thermal, Event, \etc.}
   \label{fig:3framework}
\end{figure}
RGB-based visual object tracking has garnered considerable research attention in recent years, and several works have achieved impressive tracking performance~\cite{danelljan2019ATOM, bhat2019DiMP, yan2021lSTARK, cui2022MixFormer, chen2022SimTrack, gao2022AiAtrack, lin2021SwinTrack, ye2022OSTrack, chen2023Seqtrack, wei2023ARTrack}.
However, the performance of RGB-based trackers often degrades in complex conditions, particularly due to the degradation in imaging quality.
This limitation is critical in real-world applications, especially in safety-sensitive scenarios like autonomous driving. Consequently, there is a growing interest in incorporating multimodal images to obtain more comprehensive information, which has also gained considerable attention in other areas such as object detection~\cite{tu2023EINet,jiang2019video,jiang2020learning} and semantic segmentation~\cite{zhuang2021PMF, azad2022medical, zhang2022ADSD, bai2022DCANet, liu2023M3AE, shin2023complementary, barbato2023continual,jiang2022prototypical}, among others.

Several previous studies~\cite{yan2021depthtrack, zhu2023SPT, zhu2020FANet, luo2023MACFT, zhu2023CEUTrack} have extended pre-trained RGB-based trackers~\cite{bhat2019DiMP, yan2021lSTARK, ye2022OSTrack} to other modalities, often using a symmetrical framework where information flow is symmetrical, and the pre-trained models are fully fine-tuned (see \cref{fig:3framework}(a)). However, full fine-tuning introduces numerous training parameters and encounters limitations due to the scarcity of multimodal data, resulting in a lack of generalized representation.
As an alternative, some methods~\cite{yang2022ProTrack, zhu2023ViPT} employ an asymmetrical framework and utilize prompt tuning to transfer pre-trained RGB-based trackers to other modalities (see \cref{fig:3framework}(b)), which is a parameter-efficient manner to overcome the multimodal data scarcity limitation.

\begin{figure}[t]
  \centering
   \includegraphics[width=1.0\linewidth]{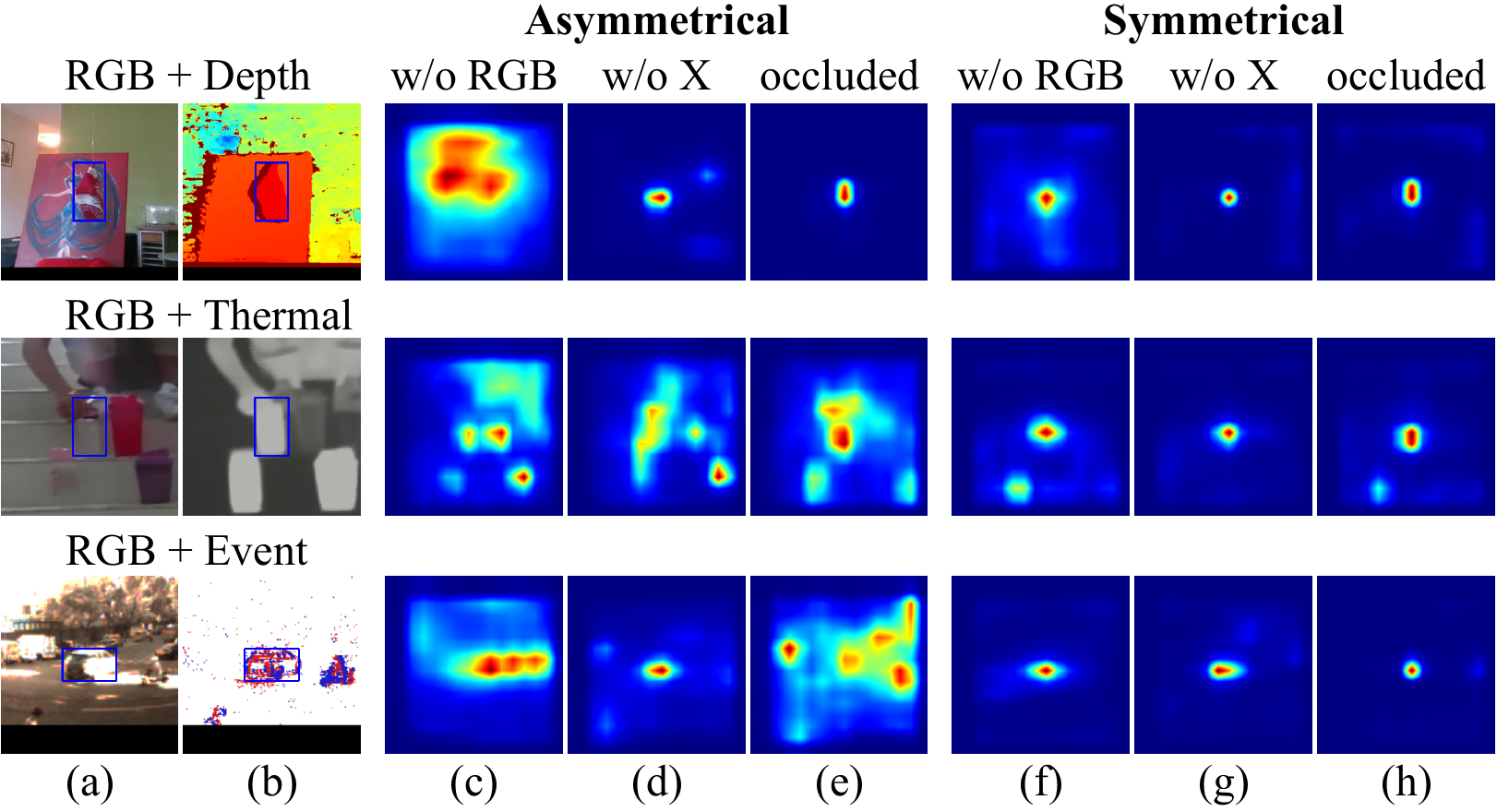}
   \vspace{-1.5em}
   \caption{\textbf{Input modality dependency comparison of multimodal object trackers}. (a)-(b) Ground truth of RGB flow and X-modal flow. (c)-(e) Score maps of ViPT~\cite{zhu2023ViPT} under RGB drop, X drop, and multimodal random occlusion conditions. (f)-(h) Score maps of our SDSTrack under RGB drop, X drop, and multimodal random occlusion conditions.
   }
   \vspace{-1em}
   \label{fig:Example}
\end{figure}

In these asymmetrical trackers, the dominance of the RGB modality persists, while other modalities serve as auxiliary modalities. 
We conducted a toy experiment to investigate the performance of trackers utilizing different frameworks in extreme scenarios such as modalities drop and modalities occlusion. As shown in \cref{fig:Example}, the results indicate a significant decrease in the performance of asymmetric trackers when the RGB modality is dropped, a moderate decrease when other modalities are dropped, and a noticeable decrease when both modalities are occluded. In contrast, symmetric trackers exhibit relatively robust performance even in challenging conditions. This experiment demonstrates that while the asymmetric structure reduces training costs, it heavily relies on the dominant modality. This reliance hinders the full utilization of information from other modalities, resulting in a lack of robustness. 

To address the aforementioned issues, we propose a novel method for multimodal tracking, termed \textbf{SDSTrack}. As illustrated in \cref{fig:3framework}(c), we integrate a symmetric framework with parameter-efficient fine-tuning (PEFT) to effectively explore and fuse multimodal information. We also introduce a complementary masked patch distillation strategy based on self-distillation learning to enhance robustness and accuracy.
Specifically, the core concept of PEFT involves freezing the pre-trained model and training only a limited number of additional parameters. This approach ensures the effective transfer of the original pre-trained model to other domains while minimizing training costs.
Building on this, we employ the idea of adapter-based tuning to transfer the feature extraction capability of RGB-based trackers to other domains and effectively fuse multimodal features. Moreover, we adopt a symmetrical structure, \ie, there is no precedence among modalities, which prevents the tracker from relying excessively on a specific modality.
Furthermore, we design a complementary masked patch distillation strategy by adding random complementary masks to the original patches during training, creating two paths. These paths share a network and perform self-distillation, enabling the model to explore information within multimodal images and improve its tracking capability, even under extreme conditions.

In summary, we make the following contributions:

$\bullet$ We propose a novel method called \textbf{SDSTrack} with a symmetrical framework. We use lightweight multimodal adaptation for parameter-efficient fine-tuning to transfer the feature extraction capability of the pre-trained model to other modalities and fuse multimodal features effectively.

$\bullet$ SDSTrack applies a complementary masked patch distillation strategy based on self-distillation learning to improve the robustness and accuracy of tracking results under extreme conditions. 

$\bullet$ Extensive experiments on several benchmarks for various modality combinations, such as RGB+Depth, RGB+Thermal, and RGB+Event, show our SDSTrack achieves state-of-the-art performance and impressive tracking results in extreme conditions.
\section{Related Works}
\label{sec:2_RelatedWorks}
\begin{figure*}[t]
  \centering
   \includegraphics[width=1.0\linewidth]{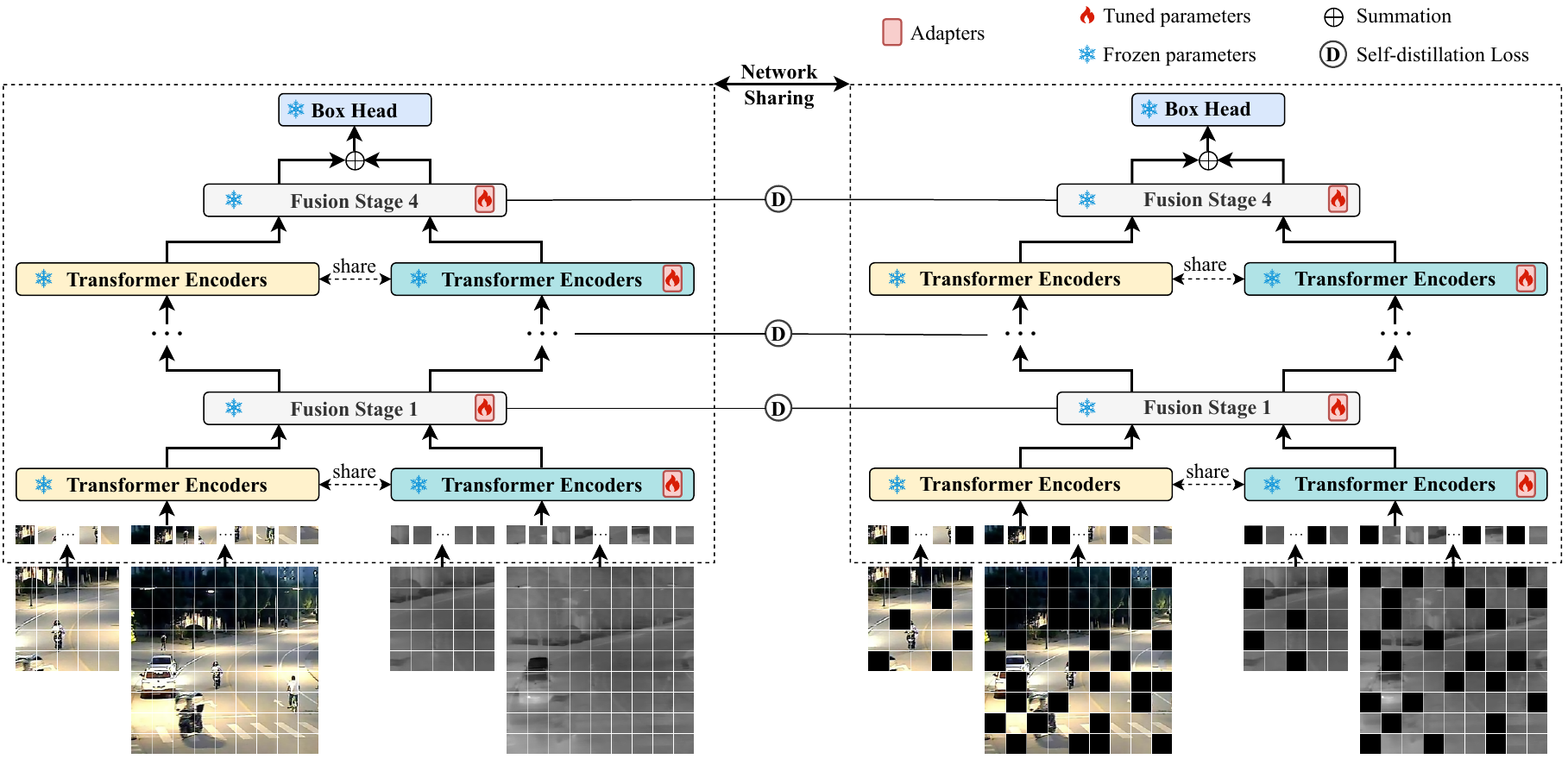}
   \caption{\textbf{The Overall pipeline of our SDSTrack}. Firstly, we apply a random masking technique to the RGB-X patch embeddings in a complementary manner, ensuring that at least one modality remains valid. Next, the masked and clean data undergo four feature extraction and fusion stages. These stages have a symmetrical structure comprising ViT blocks and adapters. In each stage, the fused features from both the masked and clean paths undergo self-distillation, which improves the accuracy and robustness of the model. Finally, the RGB and X features are combined and forwarded to the head network to obtain the prediction results.}
   \label{fig:Pipeline}
   \vspace{-1em}
\end{figure*}
\subsection{Multimodal object tracking}
Visual object tracking(VOT) has gained substantial research attention in the past decade and has found applications in various fields such as autonomous driving~\cite{fang20203d, hui20213d}, mobile robotics~\cite{wang2017real, wang2018accurate}, video surveillance~\cite{henriques2014high, li2018high, li2019siamrpn++}, human-computer interaction~\cite{yang2020grounding, zhou2023joint}, \etc. 
The goal of VOT is to track a specific target object, \ie, given a target bounding box location in the first frame, the tracker must recognize and locate this target in subsequent frames.

In recent years, due to the unstable reliability of RGB-only data in challenging scenarios, more and more studies are exploring the potential of multimodal data to address this issue. For example, CMX~\cite{zhang2023cmx} and zhang \etal~\cite{zhang2023delivering} have utilized different modalities such as RGB, Lidar, Depth, Event, Thermal, \etc for semantic segmentation. 
Similarly, in visual object tracking, researchers have dedicated considerable effort to fusing multimodal images. For example, APFNet~\cite{xiao2022APFNet} proposes an attribute-based progressive fusion network that addresses five typical challenges
by designing different branches to extract corresponding features and then fusing them. MACFT~\cite{luo2023MACFT} introduces a fusion method based on a mixed attention mechanism and uses the KL divergence loss function to enforce consistency between modal features. CEUTrack~\cite{zhu2023CEUTrack} proposes a cross-modal masking modeling strategy to facilitate modal interaction and avoid attention mechanisms from aggregating homogeneous modal information.
In addition, some works handle combinations of different modalities using a single model. For example, ProTrack~\cite{yang2022ProTrack} fuses two modalities using a weighted sum, allowing the model to adapt to RGB+depth/thermal/event combinations. However, it lacks in-depth multimodal exploration. Similarly, ViPT~\cite{zhu2023ViPT} proposes a multimodal prompting method. However, its asymmetric design does not fully exploit information from other modalities, leading to a lack of robustness.
\subsection{Parameter-efficient Fine-tuning for Vision Models}
With the development of large models~\cite{radford2021CLIP, yuan2021florence, he2022MAE}, parameter-efficient fine-tuning, \ie, PEFT, has gained increasing importance. PEFT is initially employed in natural language processing~\cite{houlsby2019parameter, lester2021power, hu2021lora, zaken2021bitfit} and has recently been widely explored in computer vision. Its core concept is to keep the pre-trained model frozen and only train a few additional parameters. This approach is particularly useful when there is limited data available for fine-tuning.
Popular PEFT methods include adapter-based tuning and prompt tuning. Adapter-based tuning~\cite{houlsby2019parameter} involves inserting trainable adapters into pre-trained models. On the other hand, prompt tuning~\cite{lester2021power} fine-tunes the model by introducing learnable prompt tokens.
In the field of multimodal visual object tracking (VOT), where multimodal tracking data is scarce, full fine-tuning is prone to overfitting and restricts the learning of generalized multimodal representations. Therefore, PEFT holds great potential in addressing these issues. However, there is limited research specifically focused on multimodal VOT using PEFT.
ProTrack~\cite{yang2022ProTrack} introduces the concept of prompt in multimodal VOT but does not apply it practically. ViPT~\cite{zhu2023ViPT} employs prompt tuning in multimodal VOT but lacks effective knowledge
recall from pre-trained models due to modality gap and assigns different importance levels to different modalities, which can lead to dependence on the primary modality. Currently, there is no research in this field based on adapter-based tuning, which is a pluggable knowledge transfer method.
\section{Method}
\label{sec:3_Method}
In this paper, we propose a novel method called SDSTrack for multimodal visual object tracking. The overall pipeline is illustrated in \cref{fig:Pipeline}. 
By training lightweight adapters, SDSTrack efficiently transfers the feature extraction capability of pre-trained RGB-based trackers to other modalities and fuses multimodal features effectively. SDSTrack can enhance robustness and accuracy through complementary masked patch distillation, even in extreme conditions.
In the following, we use X to refer to modalities other than RGB, which can be Depth, Thermal, Event, \etc.
\subsection{Symmetric Multimodal Adaptation (SMA)}
\label{Symmetric Multimodal Adaptation (SMA)}
To reduce the number of trainable parameters, mitigate the risk of overfitting, and effectively fuse different modalities, 
we propose symmetric multimodal adaptation (SMA) for adapting pre-trained RGB-based trackers to multimodal object tracking. Inspired by the parameter-efficient fine-tuning techniques~\cite{jia2022visual, xing2023multimodal, sung2022vl}, we freeze the pre-trained model and only train the proposed SMA, which comprises cross modal adaptation and multimodal fusion adaptation.

In this paper, we choose the classic one-stream RGB-based model, \eg, OSTrack~\cite{ye2022OSTrack}, as the pre-trained model, comprising a ViT~\cite{dosovitskiy2020ViT} backbone and a prediction head. Each ViT block consists of Multi-head Self-Attention (MSA), LayerNorm (LN), MultiLayer Perceptron (MLP), and residual connections.
The input of SDSTrack consists of a pair of template and search frames, \ie, template frames
$\mathbf{z}_{\text {image }}^{\text{rgb}}, \mathbf{z}_{\text {image }}^{\text {X}} \in \mathbb{R}^{H_z \times W_z \times 3}$, 
and search frames
$\mathbf{x}_{\text {image }}^{\text{rgb}}, \mathbf{x}_{\text {image }}^{\text {X}} \in \mathbb{R}^{H_x \times W_x \times 3}$, which are then projected into patch embeddings $\hat{\mathbf{z}}_{\text {rgb}}, \hat{\mathbf{z}}_{\text {X}} \in \mathbb{R}^{N_z \times D}$ and $\hat{\mathbf{x}}_{\text {rgb}}, \hat{\mathbf{x}}_{\text {X}} \in \mathbb{R}^{N_x \times D}$ by Patch Embed Layers. 
Then, the patch embeddings $\hat{\mathbf{z}}_{\text{rgb}}$ and $\hat{\mathbf{x}}_{\text{rgb}}$ are concated as $\mathbf{H}_{\text{rgb}}^{(0)}=\left[\hat{\mathbf{z}}_{\text{rgb}} ; \hat{\mathbf{x}}_{\text{rgb}}\right] \in \mathbb{R}^{\left(N_z+N_x\right) \times D}$, $\hat{\mathbf{z}}_{\text {X}}$ and $\hat{\mathbf{x}}_{\text {X}}$ are concated as $\mathbf{H}_{\text {X}}^{(0)}=\left[\hat{\mathbf{z}}_{\text {X}} ; \hat{\mathbf{x}}_{\text {X}}\right] \in \mathbb{R}^{\left(N_z+N_x\right) \times D}$, and the computation of a ViT block can be formulated as:
\begin{equation}
  \mathbf{H}{'}_{\text{rgb}}^{(l)}=\mathbf{H}_{\text{rgb}}^{(l-1)}+\operatorname{MSA}\left(\operatorname{LN}\left(\mathbf{H}_{\text{rgb}}^{(l-1)}\right)\right)
  \label{eq:MSA}
\end{equation}
\begin{equation}
  \mathbf{H}_{\text{rgb}}^{(l)}=\mathbf{H}{'}_{\text{rgb}}^{(l)} + \operatorname{MLP}\left(\operatorname{LN}\left(\mathbf{H}{'}_{\text{rgb}}^{(l)} \right)\right)
  \label{eq:MLP}
\end{equation}
where $\mathbf{H}_{\text{rgb}}^{(l-1)}$ and $\mathbf{H}_{\text{rgb}}^{(l)}$ represent the output of the $l-1$-th and $l$-th ViT block, respectively. 
\subsubsection{Cross Modal Adaptation}
\label{Cross Modal Adaptation}
Since the pre-trained model we used is trained on RGB data, there exists a modality gap to the X-modal. Therefore, we propose Cross Modal Adaptation (CMA), which transfers the feature extraction capability from the RGB domain to the X domain. The core components of CMA are adapters. As shown in \cref{fig:Adaptation}(a), the adapter is a bottleneck architecture, which consists of two fully connected (FC) layers, a GELU~\cite{hendrycks2016GELU} activation layer, and a residual connection. 
The first FC layer (FC Down) projects the input to a lower dimension, and the second FC layer (FC Up) projects it back to the original dimension. 
As illustrated in \cref{fig:Adaptation}(b), we insert adapters into ViT block after MSA and parallel to the MLP, and the calculation process is written by:
\begin{equation}
  \mathbf{H}{'}_{\text {X}}^{(l)}=\mathbf{H}_{\text {X}}^{(l-1)}+\operatorname{Adapter}\left(\operatorname{MSA}\left(\operatorname{LN}\left(\mathbf{H}_{\text {X}}^{(l-1)}\right)\right)\right)
  \vspace{-1em}
  \label{eq:X_MSA}
\end{equation}
\begin{equation}
\begin{aligned}
  \mathbf{H}_{\text {X}}^{(l)}=  \mathbf{H}{'}_{\text {X}}^{(l)} & + \operatorname{MLP}\left(\operatorname{LN}\left(\mathbf{H}{'}_{\text {X}}^{(l)} \right)\right) \\ 
  & + r \cdot \operatorname{Adapter}\left(\operatorname{LN}\left(\mathbf{H}{'}_{\text {X}}^{(l)} \right)\right)
  \vspace{-1em}
  \label{eq:X_MLP}
\end{aligned}
\end{equation}
where $\mathbf{H}_{\text {X}}^{(l-1)}$ and $\mathbf{H}_{\text {X}}^{(l)}$ are the output of the $l-1$-th and $l$-th ViT block, $r$ is a scaling factor that regulates the influence of the adapter's output weight.
\subsubsection{Multimodal Fusion Adaptation }
Previous methods often involve customized fusion modules to achieve multimodal fusion because it is commonly believed that the feature extraction backbone cannot fuse multimodal data. However, introducing the modules can lead to more tunable parameters or limited fusion performance. 

\begin{figure}[t]
  \centering
   \includegraphics[width=1.0\linewidth]{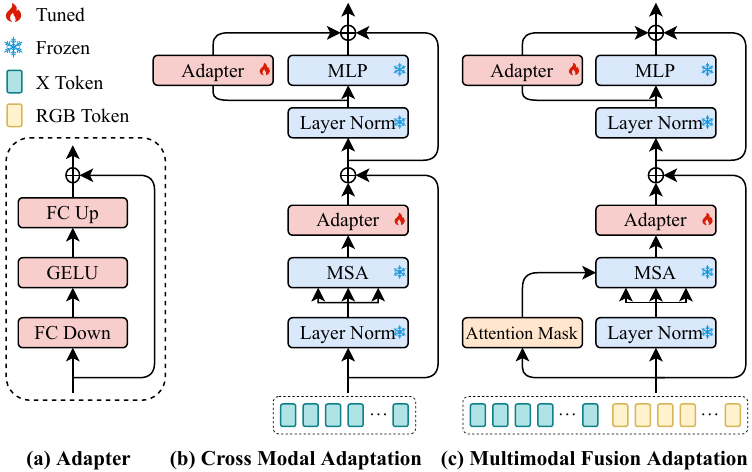}
   \caption{\textbf{The components of symmetric multimodal adaptation (SMA)}. (a) The structure of the adapter. (b) Cross Modal Adaptation for X-modal image feature extraction (c) Multimodal Fusion Adaptation for multimodal image feature fusion.}
   \vspace{-1em}
   \label{fig:Adaptation}
\end{figure}

To address this problem, we present a new strategy: \textit{reuse part of the pre-trained ViT blocks in the pre-trained model to achieve multimodal fusion}. 
Vision Transformer (ViT) processes images in a sequential modeling manner, \ie, it divides and projects images into a series of tokens, which are then processed through multiple layers of transformer models, capturing contextual dependency relationships among tokens. 
Inspired by this, we believe multimodal fusion can also be regarded as the interaction among multimodal tokens. Therefore, we propose Multimodal Fusion Adaptation (MFA), whose core components are also adapters.
Specifically, we reuse the last ViT block of each stage of encoders and integrate the adapters into the blocks as fusion stages. As shown in \cref{fig:Adaptation}(c), we insert adapters into the ViT block after MSA and parallel to the MLP. In each fusion stage, we concatenate RGB features $\mathbf{H}_{\text{rgb}}^{(l)}$ and X-modal features $\mathbf{H}_{\text {X}}^{(l)}$ as the input. Then, we feed them into components like MSA and MLP, which can learn the relationship among multimodal tokens.
To avoid overfitting and preserve the modeling capabilities of the pre-trained model, we introduce an attention mask into the MSA at each fusion stage, which sets values of RGB- and X-modal self-attention maps to zero.
The computation in the $i$-th fusion stage can be formulated as:
\begin{equation}
  \mathbf{H}{'}^{(i)}=\overline{\mathbf{H}}^{(i)}+\operatorname{Adapter}\left(\operatorname{MSA}\left(\operatorname{LN}\left(\overline{\mathbf{H}}^{(i)}\right)\right)\right)
  \vspace{-1em}
  \label{eq:fusion_MSA}
\end{equation}
\begin{equation}
\begin{aligned}
  \mathbf{H}^{(i)}=   \mathbf{H}{'}^{(i)} & + \operatorname{MLP}\left(\operatorname{LN}\left(\mathbf{H}{'}^{(i)} \right)\right) \\ 
  & + r \cdot \operatorname{Adapter}\left(\operatorname{LN}\left(\mathbf{H}{'}^{(i)} \right)\right)
  \vspace{-1em}
  \label{eq:fusion_MLP}
\end{aligned}
\end{equation}
where $\overline{\mathbf{H}}^{(i)}$ is the concatenated multimodal features to be input into $i$-th fusion stage. 
In MSA, the attention mask is:
\begin{equation}
    \mathbf{M}=\left[\begin{array}{cc}\mathbf{1} & \mathbf{0} \\ \mathbf{0} & \mathbf{1}\end{array}\right]
    \vspace{-0.5em}
    \label{eq:fusion_attn_mask}
\end{equation}
Hence, the attention map $\mathbf{A}$ after the softmax activation layer can be expressed as follows:
\begin{equation}
    \mathbf{A}=\left[\begin{array}{cc}\mathbf{0} & \mathbf{A}_{\text{RGB}, \text{X}} \\ \mathbf{A}_{\text{X}, \text{RGB}} & \mathbf{0}\end{array}\right]
    \vspace{-0.5em}
    \label{eq:fusion_attn_map}
\end{equation}
The attention mask not only prevents focusing on specific modality tokens, thus effectively facilitating multimodal interaction, but also maintains the feature extraction capability of the original pre-trained model without disturbance.
\subsection{Complementary Masked Patch Distillation}
As shown in \cref{fig:Example}, many multimodal trackers~\cite{zhu2023ViPT} rely on specific modalities and perform poorly in extreme scenarios. Additionally, these trackers fail to adequately explore the complementary information between modalities. To address these limitations, we propose a method called Complementary Masked Patch Distillation. This method aims to explore the nature of multimodal data and enhance the robustness of trackers. It involves random complementary patch mask strategy and self-distillation learning.
\subsubsection{Random Complementary Patch Mask (RCPM)}
Recently, masking strategies have found widespread application in natural language processing ~\cite{devlin2018bert, liu2019roberta} and visual applications~\cite{he2022masked, wei2022masked, xie2022simmim, shin2023complementary}.
In particular, the use of image masking strategies has become prevalent in pre-training large-capacity backbone models, allowing them to acquire generic representations for downstream tasks. One of the key principles behind masking strategies is to enable the model to reconstruct the semantic information of masked patches, thereby enhancing its understanding of the underlying data. 
Inspired by this, we employ a masking strategy to explore the complementary information between modalities, facilitating multimodal image fusion and enabling the model to avoid over-dependency on a specific modality, thus improving robustness in extreme scenarios. 
Specifically, we first obtain clean multimodal patch embeddings through image projection. Then, we randomly mask the patch embeddings of one modality, \ie, assigning their values to zero. Correspondingly, we perform random masking in the unmasked portions in the other modality. On one hand, this strategy ensures the validity of at least one modality, ensuring the accuracy of trackers. On the other hand, it enhances the model's ability to handle extreme situations, thereby improving the robustness of trackers.
\subsubsection{Self-Distillation Learning (SD)}
Merely obtaining masked embeddings through the previous RCPM can be seen as a form of data augmentation, but it is insufficient to significantly enhance the model’s robustness. We propose a self-distillation learning (SD) strategy to address this limitation. The main idea behind self-distillation learning is to leverage the clean features, which ensure the model’s accuracy, to supervise and guide the masked features, thereby improving the model’s robustness. This strategy aims to teach the model to extract complementary information from multiple modalities by referring to the clean features and adapt to extreme scenarios where modalities may be occluded or even missing.
During the training phase, we obtain both clean and masked embeddings using RCPM, resulting in two paths: the clean path and the masked path. These two paths share all the network parameters. At each fusion stage, we calculate the similarity between the fusion features from these two paths and express it using the mean squared error. This similarity calculation is used as the self-distillation loss, which encourages the model to learn to extract useful information from both the clean and masked features. The following formula expresses the self-distillation loss:
\begin{equation}
    L_{\text {SD}}=\frac{1}{s}\sum_{i=1}^{s} \operatorname{MSE}\left(\mathbf{H}^{(i)}, \widetilde{\mathbf{H}}^{(i)}\right)
\label{loss: L_SD}
\end{equation}
where $\mathbf{H}^{(i)}$ and $\widetilde{\mathbf{H}}^{(i)}$ mean the fusion features of clean data and masked data in the $i$-th fusion stage, respectively, and $\operatorname{MSE}$ is the mean squared error.
\begin{table*}[t!]
    \centering
    {
    \small
    \begin{tabular*}{\textwidth}{@{\extracolsep{\fill}} c|ccccccccccc}
        \hline
         \makebox[0.040\textwidth][c]{} & \makebox[0.040\textwidth][c]{CA3DMS} & \makebox[0.040\textwidth][c]{TSDM} & \makebox[0.040\textwidth][c]{DAL} & \makebox[0.040\textwidth][c]{ATCAIS} & \makebox[0.040\textwidth][c]{DDiMP} & \makebox[0.040\textwidth][c]{DeT} & \makebox[0.040\textwidth][c]{OSTrack} & \makebox[0.040\textwidth][c]{SPT} & \makebox[0.040\textwidth][c]{ProTrack} & \makebox[0.040\textwidth][c]{ViPT} & \makebox[0.040\textwidth][c]{\textbf{SDSTrack}} \\
         & ~\cite{liu2018ca3} & ~\cite{zhao2021TSDM} & ~\cite{qian2021DAL} & ~\cite{kristan2020ddimp} & ~\cite{kristan2020ddimp} & ~\cite{yan2021depthtrack} & ~\cite{ye2022OSTrack} & ~\cite{zhu2023SPT} & ~\cite{yang2022ProTrack} & ~\cite{zhu2023ViPT} & \textbf{(Ours)}\\
         \hline
         Pr($\uparrow$) & 0.218& 0.393& 0.512& 0.500 &0.503& 0.560& 0.536& 0.527& 0.583& 0.592& \textbf{0.619}\\
         Re($\uparrow$) &  0.228&  0.376&  0.369& 0.455 &0.469&  0.506&  0.522&  0.549&  0.573& 0.596 & \textbf{0.609}\\
         F-score($\uparrow$) &  0.223&  0.384&  0.429& 0.476 &0.485&  0.532&  0.529&  0.538&  0.578& 0.594& \textbf{0.614}\\
         \hline
    \end{tabular*}
    }
    \vspace{-0.8em}
    \caption{\textbf{Overall performance} on the DepthTrack~\cite{yan2021depthtrack} test set.}
    \vspace{-0.8em}
    \label{tab:DepthTrack_test_set}
\end{table*}
\begin{table*}[t!]
    \centering
    {
    \small
    \begin{tabular*}{\textwidth}{@{\extracolsep{\fill}} c|cccccccccc}
        \hline
         \makebox[0.062\textwidth][c]{}& \makebox[0.062\textwidth][c]{ATOM}& \makebox[0.062\textwidth][c]{DRefine}& \makebox[0.062\textwidth][c]{DMTracker}& \makebox[0.062\textwidth][c]{DeT}& \makebox[0.062\textwidth][c]{OSTrack}& \makebox[0.062\textwidth][c]{SPT} &\makebox[0.062\textwidth][c]{SBT\_RGBD}& \makebox[0.062\textwidth][c]{ProTrack}& \makebox[0.062\textwidth][c]{ViPT}& \makebox[0.062\textwidth][c]{\textbf{SDSTrack}}\\
         & ~\cite{danelljan2019ATOM} & ~\cite{kristan2021ninth} & ~\cite{kristan2022vot2022rgbd} & ~\cite{yan2021depthtrack} & ~\cite{ye2022OSTrack} & ~\cite{zhu2023SPT}  & ~\cite{kristan2022vot2022rgbd} & ~\cite{yang2022ProTrack} & ~\cite{zhu2023ViPT} & \textbf{(Ours)}\\
         \hline
         EAO($\uparrow$) & 0.505& 0.592& 0.658& 0.657& 0.676& 0.651 & 0.708 & 0.651& 0.721& \textbf{0.728}\\
         Accuracy($\uparrow$) &  0.698&  0.775&  0.758&  0.760&  0.803&  0.798 & 0.809&  0.801& \textbf{0.815}& 0.812\\
         Robustness($\uparrow$) &  0.688&  0.760&  0.851&  0.845&  0.833&  0.851 & 0.864 &  0.802& 0.871& \textbf{0.883}\\
        \hline
    \end{tabular*}
    }
    \vspace{-0.8em}
    \caption{\textbf{Overall performance} on the VOT-RGBD2022~\cite{kristan2022vot2022rgbd} dataset.}
    \vspace{-0.8em}
    \label{tab:VOT-RGBD2022}
\end{table*}
\begin{table*}[t!]
    \centering
    \small
    {
    \begin{tabular*}{\textwidth}{@{\extracolsep{\fill}} c|ccccccccccc}
        \hline
          \makebox[0.035\textwidth][c]{}&\makebox[0.035\textwidth][c]{mfDiMP} & \makebox[0.035\textwidth][c]{SGT} & \makebox[0.035\textwidth][c]{DAFNet} & \makebox[0.035\textwidth][c]{FANet} & \makebox[0.035\textwidth][c]{CAT} & \makebox[0.035\textwidth][c]{JMMAC} & \makebox[0.035\textwidth][c]{CMPP} & \makebox[0.035\textwidth][c]{APFNet} & \makebox[0.035\textwidth][c]{ProTrack} & \makebox[0.035\textwidth][c]{ViPT} & \makebox[0.035\textwidth][c]{\textbf{SDSTrack}} \\
         &  ~\cite{zhang2019mfDiMP} &  ~\cite{li2017SGT} &  ~\cite{gao2019DAFnet} &  ~\cite{zhu2020FANet} &  ~\cite{li2020CAT} &  ~\cite{zhang2021JMMAC} &  ~\cite{wang2020CMPP} &  ~\cite{xiao2022APFNet} & ~\cite{yang2022ProTrack} & ~\cite{zhu2023ViPT} &\textbf{(Ours)}\\
         \hline
         MPR($\uparrow$) &  0.646&  0.720&  0.796&  0.787&  0.804&  0.790&  0.823&  0.827& 0.795& 0.835&\textbf{0.848}\\
         MSR($\uparrow$) &  0.428&  0.472&  0.544&  0.553&  0.561&  0.573&  0.575&  0.579& 0.599& 0.617&\textbf{0.625}\\
         \hline
    \end{tabular*}}
    \vspace{-0.8em}
    \caption{\textbf{Overall performance} on the RGBT234~\cite{li2019rgbt234} dataset.}
    \vspace{-0.8em}
    \label{tab:RGBT234_test_set}
\end{table*}
\subsection{Prediction Head and Supervised Loss}
\label{Head and Loss}
We utilize and freeze the same head as OSTrack~\cite{ye2022OSTrack}, a full convolutional network(FCN) consisting of $L$ stacked Conv-BN-ReLU layers. To avoid dependence on specific modality, we take the summation of the RGB- and X-search features as the input for the head, which is different from other methods~\cite{yang2022ProTrack, zhu2023ViPT, ye2022OSTrack} that only use RGB information.
During training, SDSTrack employs the same loss function as OSTrack~\cite{ye2022OSTrack} to supervise clean and masked data tracking. 
The loss functions are represented as follows:
\begin{equation}
    L_{\text {CLEAN}}=L_{\text {cls }}+\lambda_{\text {iou }} L_{\text {iou }}+\lambda_{L_1} L_1
    \vspace{-1em}
\end{equation}
\begin{equation}
    L_{\text {MASK}}=L_{\text {cls }}+\lambda_{\text {iou }} L_{\text {iou }}+\lambda_{L_1} L_1
\end{equation}
where $\lambda_{iou}=2$ and $\lambda_{L_1}=5$ are the regularization parameters as in ~\cite{yan2021lSTARK}. Finally, the overall loss function is:
\begin{equation}
    L_{\text {track}}=L_{\text {CLEAN}} + \alpha L_{\text {MASK}} + \beta L_{\text {SD}}
\end{equation}
where $\alpha=0.3$ and $\beta=0.25$ are hyperparameters for balancing different loss functions.
\section{Experiments}
\label{sec:4_Experiments}
In this section, we first describe the experiment implementation details of the proposed SDSTrack. We then compare SDSTrack with other state-of-the-art methods on multiple benchmark datasets. Our tracker is evaluated on Depthtrack~\cite{yan2021depthtrack} and VOT22RGBD~\cite{kristan2022vot2022rgbd} for RGB-D tracking, LasHeR~\cite{li2021lasher} and RGBT234~\cite{li2019rgbt234} for RGB-T tracking, and VisEvent~\cite{wang2023visevent} for RGB-E tracking.
Lastly, we conduct robustness performance and ablation studies.
\subsection{Implementation Details}
\label{Implementation Details}
\quad\textbf{Training.} 
Our proposed SDSTrack can apply to various modality combinations. We utilize the training sets of DepthTrack~\cite{yan2021depthtrack} for RGB-D tracking, LasHeR~\cite{li2021lasher} for RGB-T tracking, and VisEvent~\cite{wang2023visevent} for RGB-E tracking.
The models are trained on 4 NVIDIA 3090Ti GPUs with a global batch size of 64. We utilize the AdamW~\cite{loshchilov2017AdamW} optimizer, with a weight decay set to $10^{-4}$. The learning rate is set to $3\times10^{-5}$ for RGB-D and $6\times10^{-5}$ for RGB-T and RGB-E. The models are trained for 15, 40, and 50 epochs for RGB-D, RGB-T, and RGB-E, respectively. Each epoch involves sampling 60k samples.

\textbf{Inference.} 
During inference,
the data follows the path of the clean data without involving the RCPM module. The tracking speed, tested on an NVIDIA 3090Ti GPU, is approximately 20.86 frames per second (\textit{fps}).
\subsection{Comparison with State-of-the-arts}
\label{Comparison with State-of-the-arts}

\quad\textbf{DepthTrack.} 
DepthTrack~\cite{yan2021depthtrack} is a comprehensive RGB-D tracking benchmark that addresses numerous challenges. It consists of 150 sequences for training and 50 sequences for testing. The evaluation metrics include precision (Pr), recall (Re), and F-score, which is calculated as $F = \frac{2Re\times Pr}{Re+Pr}$. As shown in \cref{tab:DepthTrack_test_set}, our SDSTrack achieves new state-of-the-art performance, surpassing previous SOTA 2.7\% in precision, 1.3\% in recall, and 2.0\% in F-score. 

\textbf{VOT-RGBD2022.} 
VOT-RGBD2022~\cite{kristan2022vot2022rgbd} is the most recent dataset in RGB-D tracking, comprising 127 short-term RGB-D sequences. 
The performance metric includes Accuracy, Robustness, and Expected Average Overlap (EAO). As illustrated in ~\cref{tab:VOT-RGBD2022}, our proposed SDSTrack obtains improved robustness while maintaining good precision, with a 1.2\% improvement in Robustness compared to the previous state-of-the-art performance.

\textbf{RGBT234.} 
RGBT234~\cite{li2019rgbt234} is a large-scale RGB-T tracking dataset that contains 234 pairs of videos. 
MSR and MPR are adopted for performance evaluation. As shown in \cref{tab:RGBT234_test_set}, our SDSTrack obtains new SOTA 84.8\% in MPR and 62.5\% in MSR.
\begin{figure}[t]
  \centering
   \includegraphics[width=1.0\linewidth]{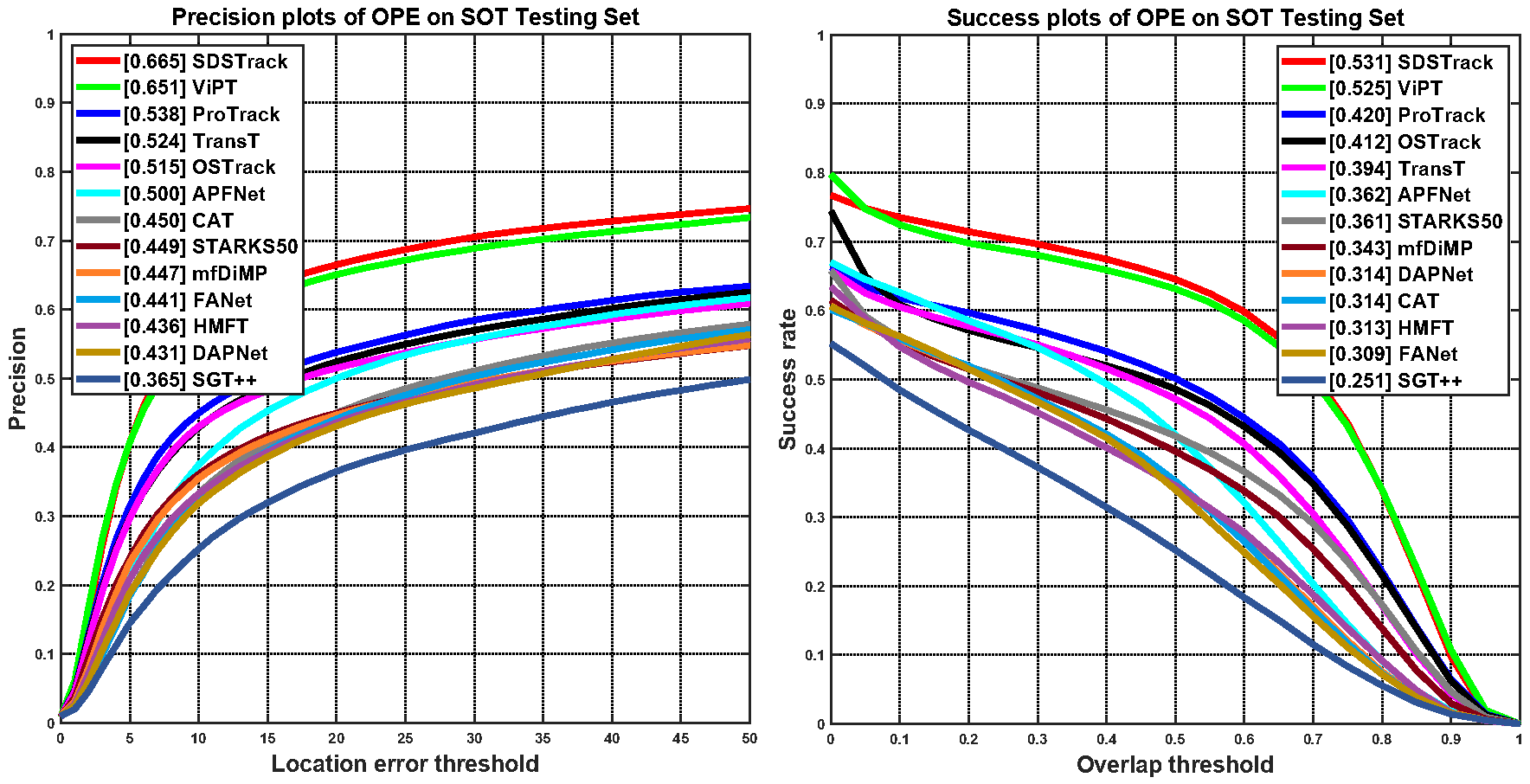}
   \caption{\textbf{Overall performance} on the LasHeR~\cite{li2021lasher} test set.}
   \label{fig:LasHeR_test_set}
   \vspace{-1em}
\end{figure}

\textbf{LasHeR.} 
LasHeR~\cite{li2021lasher} is a large-scale RGB-T dataset that comprises 979 video pairs for training and 245 pairs for testing. As shown in ~\cref{fig:LasHeR_test_set}, our SDStrack surpasses all previous SOTA trackers, obtaining the top performance of 66.5\% and 53.1\% in precision and success, respectively, where precision exceeds the previous best result by 1.4\%.

\begin{figure}[t!]
  \centering
   \includegraphics[width=1.0\linewidth]{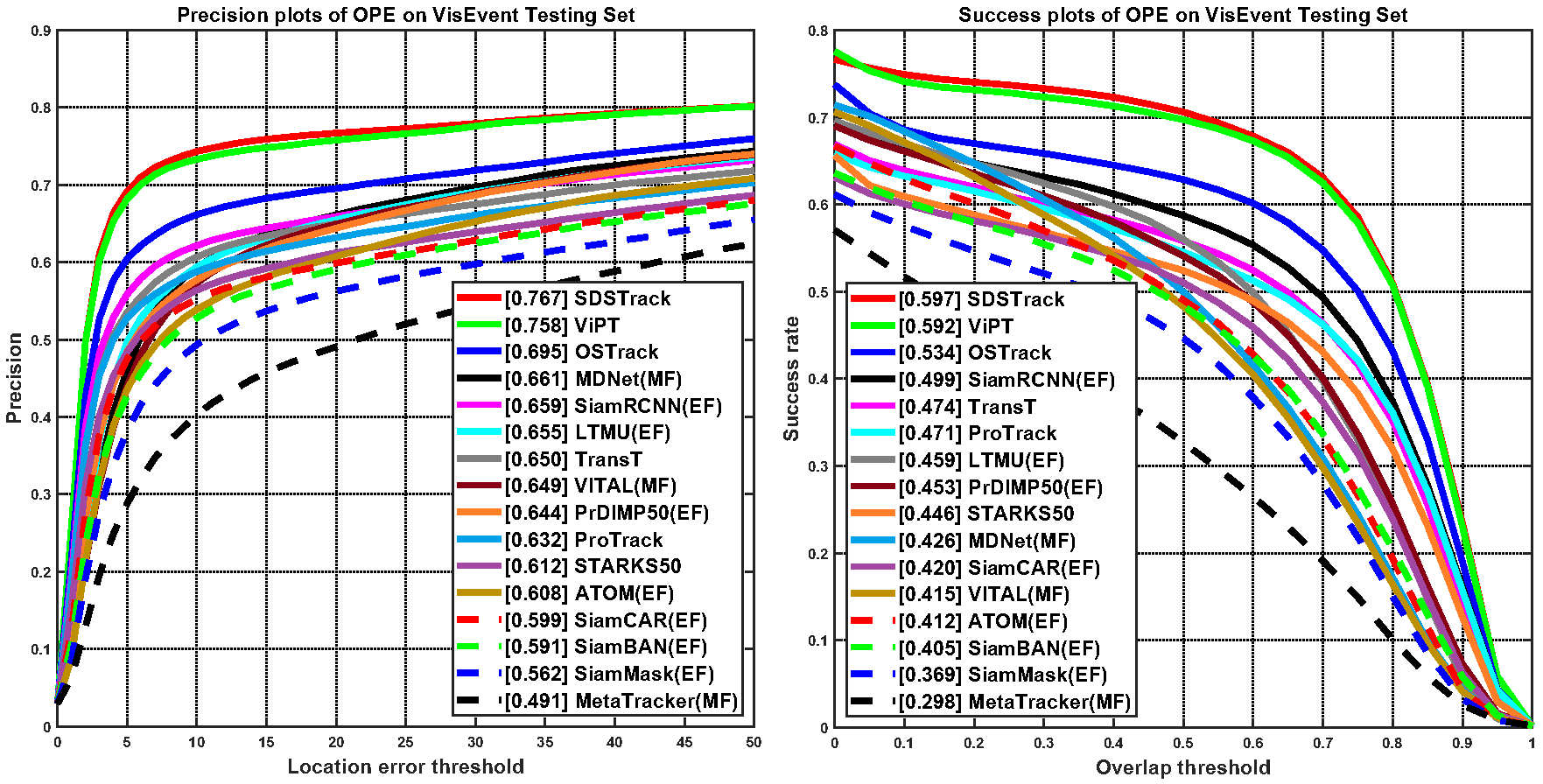}
   \caption{\textbf{Overall performance} on the VisEvent~\cite{wang2023visevent} test set.}
   \label{fig:VisEvent_test_set}
\end{figure}

\textbf{VisEvent.} 
VisEvent~\cite{wang2023visevent} is the largest dataset for RGB-E tracking. It contains 500 pairs of videos for training and 320 pairs of videos for testing. 
We compare our proposed SDSTrack with existing state-of-the-art RGB-E trackers. As reported in \cref{fig:VisEvent_test_set}, our SDStrack obtains new state-of-the-art precision and success of 76.7\% and 59.7\%, respectively.

\subsection{Exploration Study and Analysis}
\label{Exploration Study}

\quad\textbf{Robustness performance.}
To analyze the tracking performance of our method in extreme scenarios, we conducted experiments involving different scenarios, namely: (1) Dropping RGB frames with a 50\% probability. (2) Dropping X-modal frames with a 50\% probability. (3) Randomly occluding multimodal images by setting certain pixels to pure black.
As shown in ~\cref{tab:robustness}, our SDSTrack demonstrates superior robustness compared to ViPT~\cite{zhu2023ViPT}, particularly in scenarios where the RGB modality is lost. Specifically, when RGB drops, SDSTrack achieves significant improvements of 3.3\% in F-score for RGB-D tracking, 23.2\% and 16.4\% in precision and success for RGB-T tracking, 11.6\% and 7.9\% in precision and success for RGB-E tracking. Therefore, SDSTrack reduces reliance on specific modalities to a certain extent, making it effectively applicable to more challenging scenarios.
\begin{table}[h]
    \centering
    \resizebox{\columnwidth}{!}{
    \Huge
    \begin{tabular}{c|c|ccc|cc|cc}
    \hline
          \multirow{2}{*}{challenge}&\multirow{2}{*}{tracker}&\multicolumn{3}{c|}{DepthTrack~\cite{yan2021depthtrack}} &  \multicolumn{2}{c|}{LasHeR~\cite{li2021lasher}}& \multicolumn{2}{c}{VisEvent~\cite{wang2023visevent}}\\
          \cline{3-9}
          &&Pr&Re&F-score &  Pre&Suc& Pre&Suc\\ \hline
    \multirow{2}{*}{w/o RGB}&ViPT~\cite{zhu2023ViPT}&0.361&0.318&0.338&  0.306&0.268& 0.473&0.348\\
     &Ours&\textbf{0.396}&\textbf{0.340}&\textbf{0.365}&  \textbf{0.538}&\textbf{0.432}& \textbf{0.589}&\textbf{0.427}\\
    \hline
    \multirow{2}{*}{w/o X}&ViPT~\cite{zhu2023ViPT}&0.572&0.570&0.571&  \textbf{0.553}&\textbf{0.451}& 0.723&0.562\\
     &Ours&\textbf{0.587}&\textbf{0.577}&\textbf{0.582}&  0.552&0.448& \textbf{0.741}&\textbf{0.574}\\ \hline
    \multirow{2}{*}{occlusion}&ViPT~\cite{zhu2023ViPT}&0.494&0.473&0.483&  0.414&0.340& 0.627&0.439\\
      &Ours&\textbf{0.514}&\textbf{0.497}&\textbf{0.505}&  \textbf{0.545}&\textbf{0.421}& \textbf{0.639}&\textbf{0.448}\\
      \hline
    \end{tabular}}
    \caption{\textbf{The performance comparison in challenging conditions} on DepthTrack~\cite{yan2021depthtrack}, LasHeR~\cite{li2021lasher} and VisEvent~\cite{wang2023visevent}.}
    \vspace{-1em}
    \label{tab:robustness}
\end{table}

\begin{figure}[t]
  \centering
   \includegraphics[width=1.0\linewidth]{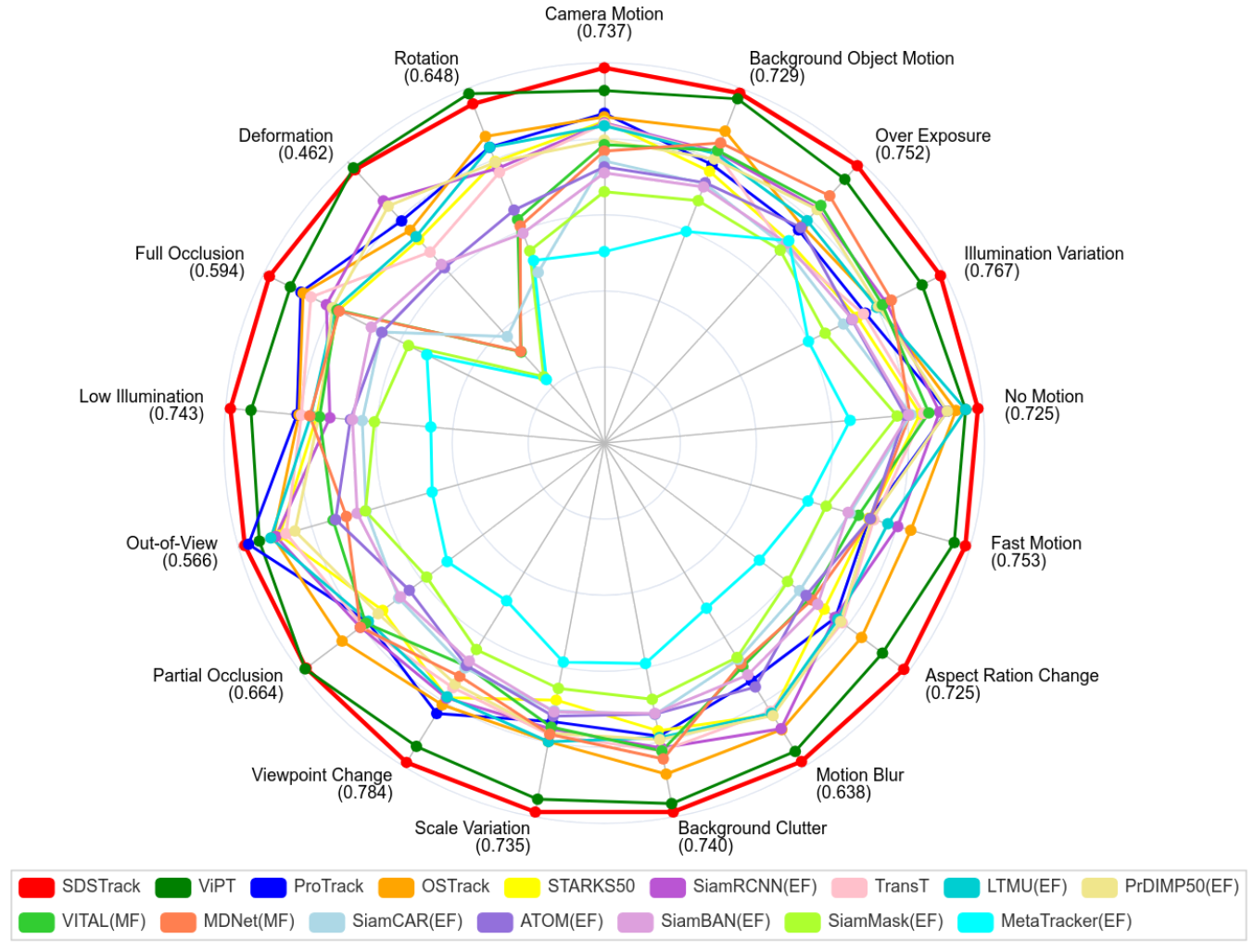}
   \caption{\textbf{Precision scores of different attributes} on the VisEvent~\cite{wang2023visevent} test set.}
   \vspace{-1em}
   \label{fig:Visevent_Precision_radar}
\end{figure}

Furthermore, we perform analysis of various challenging attributes, such as illumination variation, motion blur, out-of-view, \etc. As shown in \cref{fig:Visevent_Precision_radar}, our SDSTrack also achieves the best tracking performance in these extreme scenarios, demonstrating improved robustness. Please refer to the supplementary materials for more details.

\textbf{Effectiveness of proposed components.}
We analyze the effect of the proposed symmetric multimodal adaptation (SMA), random complementary patch mask (RCPM), and self-distillation learning (SD). Our baseline is a symmetrical structure without the mask path and SMA. It reuses the 3rd, 6th, 9th, and 11th ViT blocks as the fusion stages. The RGB and X features of the last fusion stage are added for feeding into the prediction head, where only the LN before the head and Patch Embed Layers are tuned. As shown in \cref{tab:Ablation}, the baseline equipped with SMA (Variant 1) achieves significant improvement, surpassing the baseline by 1.9\% in F-score for DepthTrack, 4.8\% and 3.0\% in precision and success for RGBT234, and 3.2\% and 3.3\% in precision and success for VisEvent. Further adding the RCPM (Variant 2) is also effective, especially improving the F-score by 1.9\% for DepthTrack. Finally, SD (Variant 3) improves 2.0\% in F-score for DepthTrack, 0.6\% and 1.0\% in precision and success for RGBT234, and 0.2\% in precision and success for VisEvent.
\begin{figure}[t]
  \centering
   \includegraphics[width=1.0\linewidth]{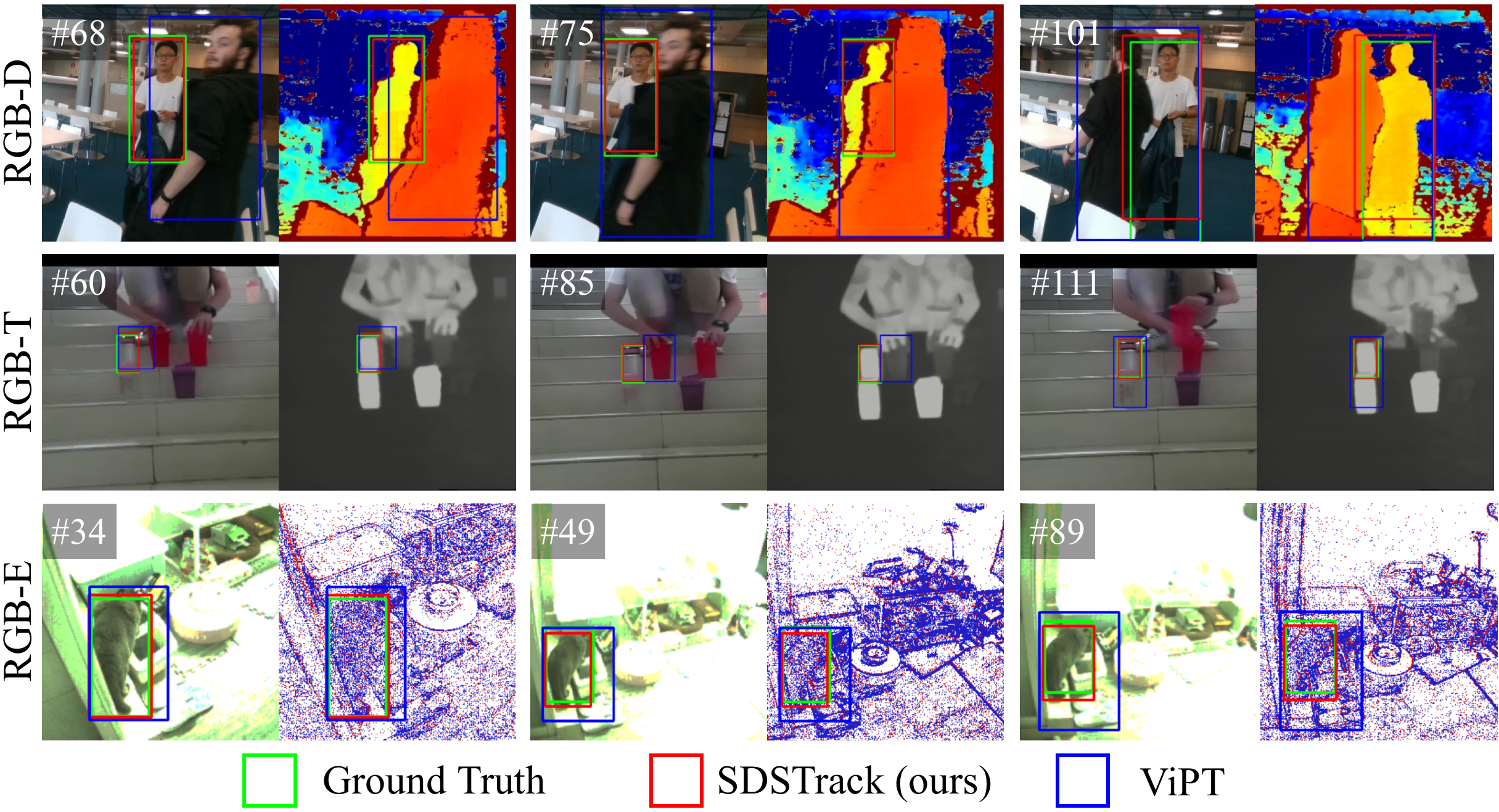}
   \caption{\textbf{Visualization results} of RGB-D, RGB-T and RGB-E.}
   \label{fig:Visualization}
\end{figure}
\begin{table}[t]
    \centering
    \resizebox{\columnwidth}{!}{
    \Huge
    \begin{tabular}{c|ccc|ccc|cc|cc}
        \hline
          \multirow{2}{*}{Variants} &\multirow{2}{*}{SMA}&\multirow{2}{*}{RCPM}&\multirow{2}{*}{SD}&\multicolumn{3}{c|}{DepthTrack~\cite{yan2021depthtrack}} &  \multicolumn{2}{c|}{RGBT234~\cite{li2019rgbt234}}& \multicolumn{2}{c}{VisEvent~\cite{wang2023visevent}}\\
          \cline{5-11}
           & & &&Pr&Re&F-score &  Pre&Suc& Pre&Suc\\ 
           \hline
    baseline &&&&0.553&0.556&0.554&  0.799&0.592& 0.728&0.562\\
     1 &$\checkmark$&&&0.578&0.569&0.573&  0.847&0.622& 0.760&0.595\\
     2&$\checkmark$&$\checkmark$&&0.599& 0.588&0.594&  0.842&0.615& 0.765&0.595\\
     3&$\checkmark$&$\checkmark$&$\checkmark$&0.619&0.609&0.614&  0.848&0.625& 0.767&0.597\\
     \hline
    \end{tabular}}
    \caption{\textbf{Ablation studies} on the effect of the components.
SMA, RCPM, and SD denote symmetric multimodal adaptation, random complementary patch mask, and self-distillation learning.}
    \vspace{-1em}
    \label{tab:Ablation}
\end{table}

\textbf{Effectiveness of the Adaptation.}
To enable a comprehensive comparison with our adaptation, we employed two training methods, freezing and fully fine-tuning (FFT), on the baseline model. The results of these comparisons are presented in \cref{tab:Ablation_Adaptation}. Remarkably, the baseline equipped with CMA surpasses the frozen- and FFT- methods by introducing only a few training parameters. Furthermore, we observed that only introducing MFA also improves performance significantly. Using both CMA and MFA contributes to the best performance.
\begin{table}[h]
    \centering
    \resizebox{\columnwidth}{!}{
    \huge
    \begin{tabular}{c|cc|ccc|cc|cc}
        \hline
          \multirow{2}{*}{Method} &Model&Tuned&\multicolumn{3}{c|}{DepthTrack~\cite{yan2021depthtrack}} &  \multicolumn{2}{c|}{LasHeR~\cite{li2021lasher}}& \multicolumn{2}{c}{VisEvent~\cite{wang2023visevent}}\\
          \cline{4-10}
           & Param (M)& Param (M)&Pr&Re&F-score &  Pre&Suc& Pre&Suc\\ 
           \hline
    Frozen&93.60&0.59&0.524&0.521&0.523&  0.492&0.398& 0.709&0.541\\
     FFT&93.60&93.60&0.570& 0.545&0.558&  0.625&0.496& 0.729&0.561\\
     \hline
     CMA&100.70&7.69&0.569&0.558&0.563&  0.632&0.506& 0.746&0.578\\
     MFA & 100.70& 7.69& 0.618& 0.599& 0.608& 0.650 & 0.520 & 0.756&0.590 \\
     CMA + MFA & 107.80& 14.79& 0.619& 0.609& 0.614& 0.665& 0.531& 0.767&0.597\\
     \hline
    \end{tabular}}
    \caption{\textbf{Ablation studies on the effect of the Adaptation}. CMA and MFA denote cross modal adaptation and multimodal fusion adaptation in Symmetric Multimodal Adaptation (SMA).}
    \label{tab:Ablation_Adaptation}
\end{table}

\textbf{Effectiveness of components on robustness.}
We analyze the components' effectiveness on the model's robustness, and the results are presented in \cref{tab:Ablation_of_robustness}. 
We find that SMA effectively enhances the robustness of the model while using RCPM alone provides a limited improvement, which can be considered a form of data augmentation. However, when RCPM is combined with SD, the model’s robustness is significantly improved.
\begin{table}[ht]
    \centering
    \resizebox{\columnwidth}{!}{
    \LARGE
    \begin{tabular}{c|ccc|cc|cc|cc}
        \hline
          \multirow{2}{*}{Variants} &\multirow{2}{*}{SMA}&\multirow{2}{*}{RCPM}&\multirow{2}{*}{SD}&\multicolumn{2}{c|}{w/o RGB}&  \multicolumn{2}{c|}{w/o X}& \multicolumn{2}{c}{occlusion}\\
          \cline{5-10}
           & & &&Pre&Suc&  Pre&Suc& Pre&Suc\\ 
           \hline
    0& $\times$ & $\times$ & $\times$ &0.298&0.251&  0.313&0.269& 0.435&0.357\\
     1 &$\checkmark$&$\times$ &$\times$ &0.472&0.389&  0.525&0.410& 0.512&0.399\\
     2&$\checkmark$&$\checkmark$&$\times$ &0.481& 0.390&  0.529&0.423& 0.523&0.415\\
     3&$\checkmark$&$\checkmark$&$\checkmark$&0.538&0.432&  0.552&0.448 & 0.545&0.421 \\
     \hline
    \end{tabular}}
    \caption{\textbf{Ablation studies on the effect of various components on robustness} on the LasHeR~\cite{li2021lasher} testing set. SMA, RCPM, and SD denote symmetric multimodal adaptation, random complementary patch mask, and self-distillation learning.}
    \label{tab:Ablation_of_robustness}
\end{table}

\textbf{Comparison on inference speed.}
We compare the inference speed of SDSTrack with previous methods. As shown in \cref{tab:inference_speed}, our method enables real-time tracking (20.86 \textit{fps}) while achieving SOTA accuracy and robustness.
\begin{table}[h]
    \centering
    \resizebox{\columnwidth}{!}
    {
    \Huge
    \begin{tabular}{c|ccc|cc|cc|c}
        \hline
          \multirow{2}{*}{Method} &\multicolumn{3}{c|}{DepthTrack~\cite{yan2021depthtrack}}&  \multicolumn{2}{c|}{LasHeR~\cite{li2021lasher}} & \multicolumn{2}{c|}{VisEvent~\cite{wang2023visevent}}&\multirow{2}{*}{FPS}
          \\
          \cline{2-8}
           &Pr&Re&F-score&  Pre&Suc & Pre&Suc &\\
           \hline
 CLGS\_D~\cite{kristan2020ddimp}& 0.584& 0.369& 0.453& -& - & -&-&7.27\\
 DDiMP~\cite{kristan2020ddimp}& 0.503& 0.469& 0.485& -& - & -&-&4.77\\
 SPT~\cite{zhu2023SPT}& 0.527& 0.549& 0.538& -& - & -&-&\textbf{25}\\
 CAT~\cite{li2020CAT}& -& -& -& 0.450& 0.314 & -&-&20\\
 DAFNet~\cite{gao2019DAFnet}& -& -& -& 0.620& 0.458 & -&-&21\\
 SiamRCNN(EF)~\cite{SiamRCNN}& -& -& -& -& -& 0.661& 0.499&4.7\\
 LTMU(EF)~\cite{LTMU}& -& -& -& -& -& 0.655& 0.459&13\\
 ViPT~\cite{zhu2023ViPT}& 0.592& 0.596& 0.594& 0.651& 0.525 & 0.758& 0.592&24.78\\
     \hline Ours&\textbf{0.619}&\textbf{0.609}&\textbf{0.614}&  \textbf{0.665}&\textbf{0.531} & \textbf{0.767}&\textbf{0.597}&20.86\\
     \hline
    \end{tabular}}
    \caption{\textbf{Comparison of inference speed}. The FPS of previous methods is obtained from respective or related papers, except for ViPT~\cite{zhu2023ViPT} tested in an environment identical to our SDSTrack.}
    \vspace{-1em}
    \label{tab:inference_speed}
\end{table}

\textbf{Visualization results.}
We visualize multimodal tracking in ~\cref{fig:Visualization}. 
The results show that SDSTrack fully utilizes multimodal information to handle challenging situations, enabling more accurate and robust tracking.
\section{Conclusion}
\label{sec:5_Conclusion}
\quad This paper introduces a novel symmetric approach called SDSTrack for multimodal visual object tracking. This approach aims to achieve parameter-efficient fine-tuning on limited multimodal data and enhance the model's robustness in extreme scenarios.
Specifically, we utilize lightweight adapter-based tuning to symmetrically fine-tune the pre-trained RGB-based tracker, transitioning the feature extraction capability from the RGB to the X domain and achieving multimodal fusion effectively.
Furthermore, we devise a random complementary patch mask strategy to obtain masked data. The masked and clean data flows are self-distilled to enhance the model's robustness in extreme scenarios. Our SDSTrack's effectiveness is demonstrated through extensive experiments on multiple benchmarks.\\
\textbf{Limitations.} Although improving accuracy and robustness in extreme scenarios, SDSTrack introduces more computational complexity during training, primarily arising from self-distillation learning. 
This problem will be a point we need to explore further in the future.\\
\\
\textbf{Acknowledgment.} This work was supported by NSFC 62088101 Autonomous Intelligent Unmanned Systems, and Zhejiang University Advanced Perception on Robotics and Intelligent Learning Lab.

{
    \small
    \bibliographystyle{ieeenat_fullname}
    \bibliography{main}
}
\end{document}